\documentclass[times,review,10pt]{elsarticle}

\usepackage{amsmath,amssymb,amsfonts}

\usepackage{graphicx}
\usepackage{subcaption}
\usepackage{float}

\usepackage{booktabs}
\usepackage{multirow}
\usepackage{tabularx}
\usepackage{array}
\usepackage{diagbox}

\usepackage{algorithm}
\usepackage{algpseudocode}

\usepackage[dvipsnames]{xcolor}

\usepackage[numbers]{natbib}
\usepackage{hyperref}
\usepackage{comment}
\usepackage{soul}
\usepackage{tikz} 

\hypersetup{
    colorlinks=true,
    linkcolor=blue,
    citecolor=blue,
    urlcolor=blue,
    filecolor=blue
}

\usepackage{caption}
\newcolumntype{Y}{>{\raggedright\arraybackslash}X}
\newcolumntype{C}[1]{>{\centering\arraybackslash}m{#1}}

\definecolor{fdblue}{RGB}{74,111,197}
\definecolor{ldorange}{RGB}{242,120,40}
\definecolor{tealgreen}{RGB}{0,128,128}

\newcommand{\blfootnote}[1]{%
  \begingroup
  \renewcommand{\thefootnote}{}%
  \footnote{#1}%
  \addtocounter{footnote}{-1}%
  \endgroup
}

\usepackage{fontawesome5}
\usetikzlibrary{positioning,calc,arrows.meta}

\definecolor{blueA}{RGB}{20,85,190}
\definecolor{greenA}{RGB}{0,125,70}
\definecolor{softblue}{RGB}{235,246,255}
\definecolor{softgreen}{RGB}{235,250,240}

\usepackage{orcidlink}

\begin{document}
\title{Scalable extraction and visualization of multi-attribute logical and functional dependencies in tabular data}

\author[1]{Chaithra Umesh,\orcidlink{0000-0002-4400-5386}}
\author[4]{Arvind Lomrore,\orcidlink{0009-0003-0446-5841}}
\author[4]{Neethu D,\orcidlink{0009-0002-5274-9593}}
\author[1]{Kristian Seegel-Schultz,\orcidlink{0000-0002-4308-6006}}
\author[1,4]{Saptarshi Bej,\orcidlink{0000-0003-1835-6139}}
\author[1,2,3]{Olaf Wolkenhauer,\orcidlink{0000-0001-6105-2937}}

\affiliation[1]{Institute of Computer Science, University of Rostock, Germany}
\affiliation[2]{Leibniz-Institute for Food Systems Biology, Technical University of Munich, Freising, Germany}
\affiliation[3]{Stellenbosch Institute for Advanced Study, South Africa}
\affiliation[4]{School of Data Science, Indian Institute of Science Education and Research, Thiruvananthapuram, India}

\begin{abstract}
Understanding the structural relationships among attributes in tabular data is fundamental to machine learning and pattern recognition. While functional dependency (FD) discovery has been extensively studied, scalable discovery of logical dependencies (LDs), particularly as the number of attributes and dependency order increase, remains underexplored. These dependencies capture non-deterministic, condition-specific relationships among pairwise or multiple attributes. Furthermore, existing approaches do not provide a unified framework for extracting multi-attribute LDs and FDs. To address these limitations, we propose \textit{LDTool} and \textit{HLDTool} to extract and visualize multi-attribute LDs and FDs from tabular data. \textit{LDTool} extends dependency discovery beyond pairwise relationships, while \textit{HLDTool} enables scalable extraction through hypergraph-guided search-space reduction. Experiments on three simulated and eleven real-world datasets demonstrate that the proposed framework extracts meaningful LDs and FDs while improving scalability. \textit{LDTool} recovers the same FDs as existing FD discovery methods with lower runtime in high-dimensional feature spaces, whereas \textit{HLDTool} enables dependency discovery in datasets with hundreds of features. The proposed framework provides interpretable visualizations of dependency structures and supports applications in exploratory data analysis and the quantitative evaluation of synthetic tabular data.

\blfootnote{Code is available at \url{https://github.com/Chaithra-U/LDTool-and-HDTool}}

\end{abstract}

\maketitle


\section{Introduction}\label{sec:introduction}

\textbf{Inter-attribute dependencies in tabular data} describe how attributes constrain or determine one another within each individual record, capturing structural relationships between attributes that extend beyond pairwise correlations. Among these, \textbf{Functional Dependencies (FDs)} constitute a foundational concept. They formalize deterministic relationships in which one set of attributes uniquely determines another \cite{lee_information-theoretic_1987}. An FD between attributes $X$ and $Y$, denoted $X \rightarrow Y$, represents a globally deterministic relationship in which each value of $X$ uniquely determines one value of $Y$ \cite{liu_discover_2012}. Formally, for all records $r_i$ and $r_j$, $r_i[X] = r_j[X] \implies r_i[Y] = r_j[Y]$. FDs provide insight into one-to-one and many-to-one mappings and are fundamental in database normalization and schema design \cite{xu_are_2024}.

Recent studies \cite{umesh_preserving_2025, long_llm-tabflow_2025} have shown that real-world datasets also exhibit \textbf{Logical Dependencies (LDs)}, rule-like constraints that hold consistently within a domain without being strictly deterministic \cite{umesh_preserving_2025}. In contrast to FDs, an LD captures rule-based constraints that emerge only under specific subsets of the attribute domain. Formally, there exists a subset $S \subseteq X$ for which the conditional relationship $P(Y=y \mid X=x)\in\{0,1\}, \; x \in S$, holds for one or more values of $Y$. Consequently, LDs represent localized rule-based constraints rather than globally deterministic mappings. For example, in clinical datasets, there is a logical dependency between sex and pregnancy status, as biologically male individuals cannot be pregnant \cite{umesh_preserving_2025}. Together, FDs and LDs reflect a wide spectrum of deterministic and rule-based relationships that shape the internal structure of tabular datasets.

A variety of algorithms have been developed for discovering FDs from tabular data, including \textit{TANE} \cite{huhtala_tane_1999}, \textit{FUN} \cite{novelli_functional_2001}, \textit{FD\_Mine} \cite{yao_fd_mine_2002}, and \textit{FDTool} \cite{buranosky_fdtool_2019}. \textit{TANE} introduced an efficient level-wise lattice traversal strategy for FD discovery, while subsequent approaches, such as \textit{FUN} and \textit{FD\_Mine}, introduced additional strategies for pruning the dependency search space  \cite{papenbrock_functional_2015}. \textit{FDTool} provides a publicly available Python implementation based on the \textit{FD\_Mine} methodology and has been experimentally evaluated against  established FD discovery algorithms \cite{buranosky_fdtool_2019}. For these reasons, we use \textit{FDTool} as the reference method for validating the FDs extracted by our framework. Recent work has extended dependency analysis beyond FDs by introducing the $Q$-function, which quantifies inter-attribute FDs and LDs within a unified framework \cite{umesh_preserving_2025}. Yet, despite these advances, FD- and LD-based structures remain uncommon in exploratory data analysis, where practitioners continue to rely largely on correlation matrices and other pairwise summaries \cite{ghosh_comprehensive_2018}. Extending dependency discovery to multi-attribute relationships introduces computational challenges. In this work, \textbf{scalability} specifically refers to dependency extraction as the number of attributes increases and higher dependency orders, i.e., dependencies involving an increasing number of LHS attributes, are considered.

\begin{figure*}[ht]
    \centering   
    \includegraphics[width=\textwidth]{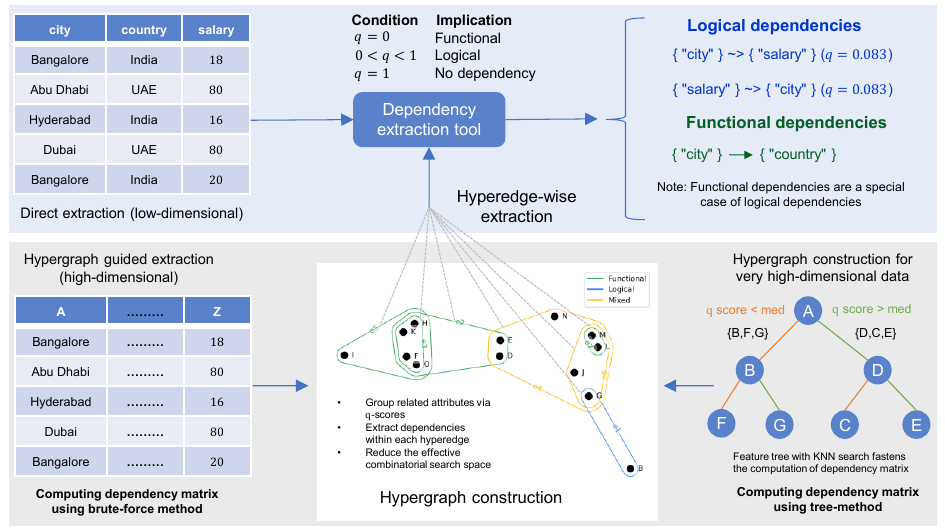}
    \caption{\textbf{Overview of the proposed dependency extraction framework.} \textit{LDTool} performs direct extraction of multi-attribute logical and functional dependencies in low-dimensional datasets. For high-dimensional feature spaces, \textit{HLDTool} employs hypergraph-guided dependency discovery, while a feature-tree–based approximation accelerates the construction of the dependency matrix for very high-dimensional datasets.}
    \label{figure: workflow}
\end{figure*}

\textbf{Research gaps:} Despite substantial progress in dependency discovery, several challenges remain in the analysis of multi-attribute dependencies in tabular data.

\textbf{Limited focus on multi-attribute LD extraction:}
Existing dependency discovery research has extensively addressed FDs and their extensions, including dependencies involving multiple attributes \cite{papenbrock_functional_2015}. While our previous work introduced the $Q$-function for quantifying pairwise LDs and FDs \cite{umesh_preserving_2025}, the extraction of LDs involving multiple attributes remains largely unexplored. To our knowledge, existing methods do not provide a unified approach for jointly extracting multi-attribute LDs and FDs from tabular data.

\textbf{Combinatorial growth of the dependency search space:}
Extracting multi-attribute dependencies requires exploring combinations of attributes, causing the search space to grow rapidly as the number of attributes and the order of dependencies increase. For $n$ attributes, lattice-based dependency discovery may involve a search space containing up to $2^n$ attribute subsets, making higher-order dependency extraction increasingly computationally expensive as feature dimensionality increases \cite{papenbrock_functional_2015}. Although existing algorithms employ pruning strategies to reduce the number of candidates evaluated, the combinatorial growth of the search space remains a challenge for high-dimensional datasets and higher-order dependency discovery.

\textbf{Difficulty in interpreting multi-attribute dependencies:}
Existing dependency discovery tools typically report extracted dependencies as individual rules, for example, through textual output files \cite{buranosky_fdtool_2019}. As the number and order of dependencies increase, interpreting the relationships among attributes from such outputs becomes increasingly difficult, particularly when dependencies involve multiple attributes. This challenge is also relevant to exploratory analysis, where effective visualization of complex relationships among multiple attributes remains important for interpreting them \cite{ghosh_comprehensive_2018}. This motivates the need for \textbf{visualization} of multi-attribute dependencies to facilitate the interpretation of complex dependency structures in tabular data.

\textbf{Our contributions:} To address these challenges, this work makes three main contributions.

\textbf{Unified LD and FD extraction framework (\textit{LDTool}):}
We develop a dependency extraction tool that jointly identifies LDs and FDs across user-specified levels of the attribute lattice. By capturing both deterministic and rule-based relationships, the framework provides a more comprehensive representation of multi-attribute dependencies than traditional FD-focused approaches.

\textbf{Scalable multi-attribute dependency discovery (\textit{HLDTool}):}
To address the exponential complexity of lattice traversal, we introduce \textit{HLDTool} that groups strongly related attributes and performs dependency discovery locally within these subsets. This substantially reduces the search space while preserving meaningful multi-attribute relationships. Additionally, we propose a VP-tree–based (Vantage Point Tree) approximation algorithm that further accelerates dependency extraction in high-dimensional feature spaces.

\textbf{Multi-view visualization of dependency structure:} We represent inter-attribute dependencies using force-directed layouts, hierarchical dendrograms, and hypergraphs. These representations reveal multi-attribute relationships within tabular data and enable interpretable exploration beyond conventional pairwise correlation analysis.

\begin{table}[ht]
\centering
\small
\setlength{\tabcolsep}{5pt}

\begin{tabular}{lrrrr}
\toprule
\textbf{Dataset} &
\textbf{Rows} &
\textbf{Attrs.} &
\textbf{FDs} &
\textbf{LDs} \\
\midrule

\href{https://github.com/Chaithra-U/LDTool-and-HDTool/blob/main/datasets/simulated_data_3.csv}{Simulated\_data\_3} \cite{Umesh2026} & 100   & 8   & 6   & 17 \\

\href{https://github.com/Chaithra-U/LDTool-and-HDTool/blob/main/datasets/simulated_data_4.csv}{Simulated\_data\_4} \cite{Umesh2026} & 100   & 15  & 33  & 214 \\

\href{https://www.kaggle.com/datasets/sahilislam007/shopping-trends-and-customer-behaviour-dataset}{Shopping behavior} \cite{islam_shopping_behavior} & 3900  & 15  & 1   & 1527 \\

\href{https://archive.ics.uci.edu/dataset/2/adult}{Adult} \cite{becker_adult_1996} & 47592 & 15  & 2   & 700 \\

\href{https://archive.ics.uci.edu/dataset/468/online+shoppers+purchasing+intention+dataset}{Online shopping} \cite{sakar_online_shoppers_2018} & 12330 & 18  & 1   & 518 \\

\href{https://codeocean.com/capsule/1269964/tree/v1/data/migraine.csv}{Migraine} \cite{sanchez_migraine_2020} & 377   & 20  & 33  & 1308 \\

\href{https://archive.ics.uci.edu/dataset/73/mushroom}{Mushroom} \cite{mushroom_1981} & 5644  & 22  & 102 & 2526 \\

\href{https://www.kaggle.com/datasets/mohankrishnathalla/global-house-purchase-decision-dataset}{Global house purchase} \cite{thalla_global_house} & 10000 & 24  & 317 & 1856 \\

\href{https://www.kaggle.com/datasets/sohaibdevv/consumer-shopping-behavior-and-preference-study-2026}{Consumer shopping trends 2026} \cite{consumer_shopping_2026} & 11789 & 25  & 382 & 9420 \\

\href{https://github.com/Chaithra-U/LDTool-and-HDTool/blob/main/datasets/simulated_data_7.csv}{Simulated\_data\_7} \cite{Umesh2026} & 5000  & 33  & 33  & 16799 \\

\href{https://www.kaggle.com/datasets/nitikachandel95/student-social-media-impact-dataset?select=global_student_digital_behavior_dataset.csv}{Student digital behavior} \cite{chandel_student_social_media} & 500   & 47  & 396 & 3519 \\

\href{https://archive.ics.uci.edu/dataset/116/us+census+data+1990}{US Census} \cite{meek_us_census_2001} & 10000 & 68  & 20  & 224767 \\

\href{https://archive.ics.uci.edu/dataset/125/insurance+company+benchmark+coil+2000}{Tic Insurance} \cite{putten_coil_2000} & 5220  & 86  & 132 & 51026\\

\href{https://www.kaggle.com/datasets/raghadalharbi/breast-cancer-gene-expression-profiles-metabric}{METABRIC} \cite{curtis_metabric_2012} & 1092  & 642 & 7   & 1512 \\

\bottomrule
\end{tabular}

\caption{Summary of the experimental datasets together with the number of extracted FDs and LDs. Dependencies are reported up to the third layer, except for \texttt{METABRIC} and \texttt{Tic insurance}, where only first and second layer dependencies are extracted because of the high feature dimensionality.}
\label{table:dataset_table}
\end{table}
 
Experiments on three simulated and eleven real-world datasets demonstrate that the proposed framework extracts meaningful multi-attribute LDs and FDs while substantially improving scalability for high-dimensional tabular data. When restricted to FD extraction, \textit{LDTool} recovers the same multi-attribute FDs as \textit{FDTool} up to three dependency layers while reducing runtime, especially for high-dimensional datasets (Refer to the Table \ref{table:time}). Additionally, multi-view visualizations consistently reveal interpretable dependency structures. Together, these results demonstrate scalable extraction and visualization of multi-attribute dependencies in tabular data.

Many real-world datasets contain higher-order dependencies that cannot be captured by pairwise analysis alone. Examples include purchasing patterns in e-commerce data and gene-expression interactions in biomedical datasets. The proposed approach enables structured discovery and visualization of such relationships through a unified extraction framework. Beyond exploratory analysis, the resulting dependency structures provide a basis for evaluating preservation of multi-attribute relationships in synthetic tabular data by comparing structural consistency between real and synthetic datasets.

\section{Related research}

Most existing research on dependency discovery in tabular data has primarily focused on FDs. Early foundational systems such as \textit{TANE} \cite{huhtala_tane_1999} and \textit{Dep-Miner} \cite{goos_efficient_2000} established efficient frameworks for discovering global FDs from relational datasets. \textit{TANE}
introduced a level-wise lattice traversal strategy using stripped partitions to efficiently validate exact and approximate dependencies, while \textit{Dep-Miner} utilized agree sets and difference sets to derive minimal FDs and Armstrong relations. \textbf{These approaches provided scalable foundations for FD mining but remained primarily focused on dependencies defined by classical FD semantics.}

Subsequent research extended classical FD discovery toward more flexible dependency formulations. Relaxed functional dependencies generalize classical FDs by relaxing different aspects of exact satisfaction, including approximation, conditional validity, similarity, and probabilistic relationships \cite{caruccio_relaxed_2016}. Conditional Functional Dependencies (CFDs), for example, represent dependencies that hold only under specific conditions or subsets of the data. Bohannon et al. \cite{bohannon_conditional_2007} introduced CFDs for data cleaning, while later works explored their use in inconsistency detection \cite{fan_conditional_2008}, rule discovery, and data quality analysis \cite{chiang_discovering_2008}. Inclusion dependencies (INDs) constitute another established dependency class and express containment relationships among attribute projections or combinations, with applications such as foreign-key discovery and schema analysis \cite{duersch_inclusion_2019}. \textbf{Although relaxed FDs, CFDs, and INDs broaden the classical dependency framework, their semantics differ from the LDs considered in this work.} Following our previous $Q$-function formulation \cite{umesh_preserving_2025}, an LD describes a directional inter-attribute relationship in which the LHS constrains the possible values of the RHS without determining it functionally, with its strength quantified through $q$-scores.

To better handle noisy, imperfect real-world datasets, approximate and statistical methods for dependency discovery were later introduced. Systems such as \textit{AIDFD} \cite{bleifus_approximate_2016}, \textit{Pyro} \cite{kruse_efficient_2018}, and statistical FD discovery frameworks \cite{zhang_statistical_2020}, relaxed strict deterministic assumptions and enabled the discovery of soft, noisy, and approximate dependencies. Information-theoretic and probabilistic approaches further expanded dependency analysis beyond exact relational constraints \cite{kinney_equitability_2014}. While these methods improved robustness to noise and scalability, they largely \textbf{focused on approximate global relationships and pairwise statistical associations rather than multi-attribute LDs.}

Graph-based representations have also been explored for modeling and visualizing dependency structures. \textit{Dependency Networks} \cite{heckerman_dependency_2013} and \textit{SWRL} dependency visualization \cite{hassanpour_visualizing_2010} approaches demonstrated that graphical representations can improve the interpretability of complex dependency relationships. These methods enabled structural exploration of dependencies using graph-based layouts and rule visualization strategies. \textbf{However, existing visualization approaches were not specifically designed for unified FD and LD discovery in tabular datasets and generally did not represent higher-order dependencies involving multiple attributes}.

Several surveys and conceptual works have further contributed to the theoretical understanding of dependency semantics, dependency discovery algorithms, and data quality constraints \cite{fan_dependencies_2008}. Research has also explored extensions such as differential dependencies \cite{song_differential_2011}, sequential dependencies \cite{golab_sequential_2009}, and approximate uniqueness constraints \cite{kruse_efficient_2018}. \textbf{Although these studies have broadened the scope of dependency discovery, they do not provide a unified framework for the scalable extraction, quantification, and visualization of multi-attribute LDs and FDs in tabular data}.

Among practical FD discovery tools, \textit{FDTool} \cite{buranosky_fdtool_2019} is a Python re-implementation of the \textit{FD\_Mine} algorithm for discovering minimal FDs and candidate keys from tabular data. \textit{FD\_Mine} uses a level-wise search strategy together with equivalence-based pruning to reduce the number of candidate dependencies that need to be evaluated. Experiments with \textit{FDTool} showed that increasing the number of attributes has a substantially greater effect on runtime and memory usage than increasing the number of rows.

More recent research has highlighted the growing importance of preserving LDs and FDs during synthetic tabular data generation. Early constraint-aware approaches such as \textit{Kamino} \cite{ge_kamino_2020} and \textit{CuTS} \cite{vero_cuts_2023} enforce predefined FDs and LDs during synthesis to improve data fidelity and semantic consistency. More recent methods have shifted toward automatically learning dependency structures from data. \textit{SAGE} \cite{yang_sage_2026} discovers inter-attribute dependencies and incorporates them into the generation process to better preserve logical consistency between real and synthetic datasets. \textit{PRISM} \cite{guan_prism_2026} further improves dependency-aware generation by explicitly modeling and preserving structural relationships among attributes during synthesis, while \textit{MoE-T} \cite{dhooghe_moet_2026} leverages a mixture-of-experts architecture to capture complex feature interactions and higher-order dependencies across heterogeneous data distributions. Similarly, \textit{LLM-TabFlow} \cite{long_llm-tabflow_2025} employs large language models to learn latent inter-column logical relationships that are preserved during diffusion-based data generation. Collectively, these studies demonstrate the increasing importance of dependency-aware synthetic data generation. \textbf{However, most existing approaches primarily rely on dependencies to guide or constrain the generation process, rather than directly discovering, quantifying, and visualizing multi-attribute LDs and FDs from tabular data}.

Despite substantial progress in FD discovery, approximate dependency mining, and dependency-aware data generation, multi-attribute LD extraction remains underexplored. Existing methods have largely been developed for specific dependency classes, including deterministic FDs, conditional dependencies, and approximate dependencies, while \textbf{methods that extract both multi-attribute LDs and FDs within a common framework are lacking.} In addition, existing dependency discovery tools typically report extracted dependencies as individual rules, making the relationships among attributes difficult to interpret as the number and order of dependencies increase.

Our work addresses these gaps by introducing a unified framework for extracting multi-attribute LDs and FDs from tabular data. The proposed framework combines scalable dependency extraction with hypergraph-guided search and provides interpretable visualizations of multi-attribute dependencies using force-directed graphs, dendrograms, and hypergraphs.
    
\section{Methodology}

\subsection{Preliminaries: the \texorpdfstring{$Q$}{Q}-function}

Let $D$ be a tabular dataset with feature set $F = \{f_1, f_2, \dots, f_n\}$. A \emph{rule} is written $S \rightarrow y$, where $S \subseteq F$ is a set of $k = |S|$ attributes forming the left-hand side (LHS) and $y \in F \setminus S$ is the right-hand side (RHS) attribute. We refer to $k=|S|$ as the \emph{dependency order} and say that the rule belongs to \emph{layer} $k$. For a rule $S \rightarrow y$, let $D[S,y]$ denote the projection of $D$ onto $S \cup \{y\}$ with duplicate rows removed. We define three quantities from this projection:

\begin{description}
    \item[{$\mathrm{dom}(S\rightarrow y)$}:]
    Projection of $D[S,y]$ onto $S$ with duplicates removed.
    Its size $|\mathrm{dom}(S\rightarrow y)|$ is the number of distinct
    LHS combinations.

    \item[{$\mathrm{img}(S\rightarrow y)$}:]
    Projection of $D[S,y]$ onto $y$ with duplicates removed.
    Its size $|\mathrm{img}(S\rightarrow y)|$ is the number of distinct
    RHS values.

    \item[{$|D[S,y]|$}:]
    Number of distinct (LHS, RHS) combinations observed together.
    This counts \emph{distinct} combinations rather than raw row counts;
    therefore, repeated rows in $D$ do not change the score.
\end{description}

The dependency strength between $S$ and $y$ is measured by the $Q$-function, introduced for the pairwise case in \cite{umesh_preserving_2025} and generalized here to multi-attribute LHS sets:
\[
q(S \rightarrow y) \;=\;
\begin{cases}
0, & |\mathrm{dom}(S\rightarrow y)| = 0 \ \text{or}\ |\mathrm{img}(S\rightarrow y)| \leq 1, \\[6pt]
\dfrac{\dfrac{|D[S,y]|}{|\mathrm{dom}(S\rightarrow y)|} - 1}{|\mathrm{img}(S\rightarrow y)| - 1}, & \text{otherwise}.
\end{cases}
\]
We write $strength(S\rightarrow y) := 1 - q(S\rightarrow y)$ for the complementary score, which we use in practice since higher values then mean stronger dependency.

\paragraph{The score is always between 0 and 1}
Each distinct LHS combination in $\mathrm{dom}(S\rightarrow y)$ is paired with at least one and at most $|\mathrm{img}(S\rightarrow y)|$ distinct RHS values. Summing over all LHS combinations gives
\[
|\mathrm{dom}(S\rightarrow y)| \;\le\; |D[S,y]| \;\le\; |\mathrm{dom}(S\rightarrow y)| \cdot |\mathrm{img}(S\rightarrow y)|.
\]
Substituting these two bounds into the formula above shows directly that $0 \le q(S\rightarrow y) \le 1$ always holds.

\paragraph{Classification of FD vs LD} The value of $q$ tells us what kind of relationship $S$ and $y$ have:
\begin{description}
    \item $q = 0$: a \emph{functional dependency (FD)} — every LHS combination determines the RHS value exactly.
    \item $0 < q < 1$: a \emph{logical dependency (LD)} —a non-deterministic but structured, condition-specific relationship.
    \item $q = 1$: no dependency.
\end{description}
Lower $q$ (higher strength) means a stronger relationship. Since an FD is simply the special case $q=0$, every FD is also an LD in this framework, and the same scoring function covers both.

\subsection{Unified multi-attribute LD and FD extraction (\textit{LDTool})} \label{method: ldtool}

Most existing dependency discovery methods primarily focus on FDs and, when they do consider LDs, they treat them merely as pairwise relationships. \textit{LDTool} removes both restrictions: it discovers LDs and FDs together, at any LHS size up to a chosen maximum layer $L$.

\textit{LDTool} takes as input the dataset $D$, the maximum layer $L$, a minimum
strength threshold $\tau_s$, an LD improvement threshold $\tau_{LD}$, and the
requested dependency type $T \in \{\mathrm{FD}, \mathrm{LD}, \mathrm{both}\}$.
It first enumerates candidate rules layer by layer and keeps those whose
strength exceeds $\tau_s$; it then removes redundant, non-minimal rules and
finally returns the type of dependency that was requested. In this work, we
use $\tau_s = 0.8$, which keeps strong LDs together with all FDs.

\begin{algorithm}
\caption{Unified multi-attribute LD and FD extraction}
\label{algorithm:LDFD extraction}
\begin{algorithmic}[1]

\Require Dataset $D$ with features $F$, maximum layer $L$, minimum strength $\tau_s$, LD improvement threshold $\tau_{LD}$, dependency type $T \in \{\mathrm{FD}, \mathrm{LD}, \mathrm{both}\}$
\Ensure Extracted dependencies $\mathcal{R}$

\State $\mathcal{R} \gets \emptyset$

\Comment{Step 1: generate candidates}
\For{each RHS attribute $y \in F$}
    \State $X \gets F \setminus \{y\}$
    \For{$k = 1$ to $\min(L, |X|)$}
        \For{each subset $S \subseteq X$ with $|S| = k$}
            \State Compute $q(S \rightarrow y)$ and $\mathrm{strength}(S \rightarrow y) = 1 - q(S \rightarrow y)$
            \If{$strength(S \rightarrow y) \ge \tau_s$}
                \State Add $(S \rightarrow y)$ to $\mathcal{R}$
            \EndIf
        \EndFor
    \EndFor
\EndFor

\Comment{Step 2: remove non-minimal rules (Definition~1)}
\For{each rule $(S \rightarrow y) \in \mathcal{R}$}
    \For{each proper subset $S' \subsetneq S$}
        \If{$(S' \rightarrow y) \in \mathcal{R}$}
            \If{$q(S \rightarrow y)=0$ and $q(S' \rightarrow y)=0$}
                \State Remove $(S \rightarrow y)$ from $\mathcal{R}$
                \Comment{exact minimality for FDs}
                \State \textbf{break}
            \ElsIf{$q(S \rightarrow y)>0$ and $\mathrm{strength}(S \rightarrow y)-strength(S' \rightarrow y) < \tau_{LD}$}
                \State Remove $(S \rightarrow y)$ from $\mathcal{R}$
                \Comment{threshold-based minimality for LDs}
                \State \textbf{break}
            \EndIf
        \EndIf
    \EndFor
\EndFor

\Comment{Step 3: return the requested type}
\If{$T = \mathrm{FD}$}
    \State \Return only FDs in $\mathcal{R}$
\ElsIf{$T = \mathrm{LD}$}
    \State \Return only LDs in $\mathcal{R}$
\Else
    \State \Return $\mathcal{R}$
\EndIf

\end{algorithmic}
\end{algorithm}

\paragraph{Definition 1 (Minimality)}
A rule $(S \rightarrow y) \in \mathcal{R}$ is \emph{minimal} unless there exists a proper subset $S' \subsetneq S$ with $(S' \rightarrow y) \in
\mathcal{R}$ such that:
\begin{itemize}
    \item If $(S \rightarrow y)$ is an FD: $(S' \rightarrow y)$ is also an FD; or
    \item If $(S \rightarrow y)$ is an LD: $\mathrm{strength}(S \rightarrow y) - strength(S' \rightarrow y) < \tau_{LD}$.
\end{itemize}
If such an $S'$ exists, $(S \rightarrow y)$ is removed. Note that $S'$ ranges over \emph{every} proper subset of $S$, not only the subsets one attribute smaller: a higher-order LD must improve on the strength of all of its smaller ancestors by at least $\tau_{LD}$, not just its immediate parent.

For FDs, this is the classical rule: a higher-order FD is redundant whenever a smaller exact FD with the same RHS already exists. For LDs, classical minimality would discard higher-order rules even when they capture genuinely new conditional structure, so we keep a higher-order LD only when it improves strength by at least $\tau_{LD}$ over every smaller LD it could be derived from. We found $\tau_{LD} = 0.2$ gives a good balance between removing redundant rules and keeping informative ones.

\begin{table}[ht]
\centering
\footnotesize
\setlength{\tabcolsep}{2.5pt}

\begin{tabular}{ll}
\toprule
\textbf{Dataset dimensionality} & \textbf{Tool} \\
\midrule

Low-dimensional  & \textit{LDTool}  \\
High-dimensional & \textit{HLDTool} \\

\bottomrule
\end{tabular}

\caption{Selection of dependency extraction tools based on dataset dimensionality.}
\label{tab:tool}

\end{table}

\paragraph{Remark (FD-only extraction)}
When only FDs are requested ($T=\mathrm{FD}$), it is not necessary to compute $q$ at all: checking whether every distinct LHS combination maps to exactly one RHS value is equivalent to testing $q=0$, and is much cheaper to compute. This equivalence carries over to minimality too: since FD minimality (Def.~1) only ever compares a candidate against \emph{other exact} dependencies, the
classical Apriori-style pruning rule: \textit{skip a candidate whose LHS is a superset of an already-confirmed FD} yields exactly the same minimal set of FDs as running Algorithm \ref{algorithm:LDFD extraction} restricted to $T=\mathrm{FD}$. We therefore use this faster exact check whenever only FDs are needed, and the general $q$-based procedure above otherwise.

After minimality filtering, the surviving dependencies are reported layer by layer, split into LDs and FDs, as shown in Figure \ref{figure:dependencies}.

\subsection{Hypergraph-guided scalable dependency discovery (\textit{HLDTool})}
\label{method: hldtool}

\textit{LDTool} searches every subset of attributes up to layer $L$, so the number of candidates grows quickly as the number of attributes increases. \textit{HLDTool}
avoids this by first grouping attributes that are strongly related, and then running \textit{LDTool} independently within each group.
We build a \emph{hypergraph} $H = (V, E)$ where $V = F$ is the set of attributes and $E$ is the set of hyperedges, each containing a group of mutually related attributes. Construction proceeds in three steps:

\begin{enumerate}
    \item \textbf{Pairwise relatedness.} For each attribute $f_i$, the pairwise $q$-scores $q(f_i \rightarrow f_j)$ are computed with respect to every other attribute $f_j$. Attributes satisfying $q(f_i \rightarrow f_j) < \tau_H$, where $\tau_H$ is a user-defined threshold, are retained as candidate attributes.

    \item \textbf{Grouping.} The candidate attributes are ranked according to their $q$-scores, and the top-$k$ candidates are selected. Together with the corresponding attribute $f_i$, they form a hyperedge. Duplicate hyperedges are removed after construction.

    \item \textbf{Size control.} The parameter $k$ controls the maximum number of candidate attributes retained in each hyperedge, providing a trade-off between dependency coverage and computational efficiency (see Figure \ref{fig:hyperedge_size}).
\end{enumerate}

Once $H$ is built, Algorithm \ref{algorithm:LDFD extraction} is run
independently inside each hyperedge instead of over the full feature set, which sharply reduces the number of candidate subsets while keeping the attribute groups that matter together.

The overall framework, therefore, has three modes, chosen by dataset size: for low-dimensional data, \textit{LDTool} searches the full feature space directly, for high-dimensional data, \textit{HLDTool} restricts the search to hyperedges, and
for very high-dimensional data, an approximate version of the pairwise $Q$-matrix (Section \ref{method: VP-tree}) is computed first to make even the hyperedge construction step affordable. Figure \ref{figure: workflow} summarises this workflow.

\subsection{Fast approximation of the \texorpdfstring{$Q$}{Q}-matrix using a vantage-point tree}
\label{method: VP-tree}

Building the pairwise $Q$-matrix used in hyperedge construction requires evaluating $\mathrm{strength}(f_i \rightarrow f_j)$ for every ordered pair of attributes, which costs $\mathcal{O}(n^2)$ evaluations for $n$ attributes. For very high-dimensional data, this becomes a bottleneck in its own right, before any dependency extraction has even started. We speed this up with an
approximate nearest-neighbor search over the attributes themselves.

\begin{algorithm}[ht]
\caption{Building the vantage-point feature tree}
\label{alg:build_tree}
\begin{algorithmic}[1]
\Require Set of feature indices ${F}$
\Ensure Root node of the vantage-point tree ${T}$
\Function{BuildVPTree}{${F}$}
    \If{$|{F}| \leq 1$}
        \State \Return \textbf{NewNode}(${F}[0]$)
    \EndIf
    \State Select random pivot $f_p \in {F}$
    \State Compute distances ${D} \leftarrow \{ d(f_p, f_x) \mid f_x \in {F} \setminus \{f_p\} \}$
    \Comment{$d(f_p,f_x) = q(f_p \rightarrow f_x)$}
    \State Sort ${D}$ to find median $\mu$
    \State $S_L \leftarrow \{ f_x \mid d(f_p, f_x) < \mu \}$
    \Comment{closer to $f_p$ = stronger dependency}
    \State $S_R \leftarrow \{ f_x \mid d(f_p, f_x) \ge \mu \}$
    \State node $\leftarrow$ \textbf{NewNode}($f_p$)
    \State node.median $\leftarrow \mu$
    \State node.left $\leftarrow$ \Call{BuildVPTree}{$S_L$}
    \State node.right $\leftarrow$ \Call{BuildVPTree}{$S_R$}
    \Return node
\EndFunction
\end{algorithmic}
\end{algorithm}

We treat each attribute as a point in a space where the distance between two attributes $f_i, f_j$ is
\[
d(f_i,f_j) := q(f_i \rightarrow f_j),
\]
so a \emph{smaller} distance means a \emph{stronger} dependency, matching the definition of $q$ directly: lower $q$, shorter distance. This distance is not symmetric, since $q(f_i\rightarrow f_j) \ne q(f_j\rightarrow f_i)$ in general, and so it does not necessarily satisfy the triangle inequality either. Because of this, the tree built below is a heuristic index rather
than an exact metric structure; we validate it empirically against brute-force computation in Table~\ref{tab:fast-tree-vs-brute-force}. For this reason, we call it a \emph{vantage-point (VP) tree} rather than a KP-tree: unlike a KD-tree, which splits points along their own coordinate axes, our tree splits attributes purely by distance from a chosen pivot attribute, which is all we
have available here.

\paragraph{Building the tree} Algorithm~\ref{alg:build_tree} builds the tree
recursively:
\begin{enumerate}
    \item A pivot attribute $f_p$ is chosen at random from the current set.
    \item The distance $d(f_p, f_x)$ is computed from $f_p$ to every other
    attribute $f_x$ in the current set.
    \item The median distance $\mu$ splits the remaining attributes into a
    closer group $S_L$ (stronger dependency on $f_p$) and a farther group
    $S_R$.
    \item The same procedure is applied recursively to $S_L$ and $S_R$.
    \item Recursion stops, and a leaf is created, once a subset contains a
    single attribute.
\end{enumerate}

\begin{algorithm}[ht]
\caption{Vantage-point tree $k$-NN dependency search}
\label{alg:knn_search}
\begin{algorithmic}[1]
\Require Vantage-point tree root $node$, Query $f_q$, Neighbors $k$
\Ensure Max-heap of nearest neighbors $H$
\State Initialize $H \leftarrow \emptyset$, $\tau \leftarrow \infty$
\Procedure{Search}{$node$}
    \If{$node$ is \textbf{null}} \Return \EndIf
    \State $dist \leftarrow d(f_q, node.feature)$
    \If{$node.feature \neq f_q$ \textbf{and} $dist < \tau$}
        \State $H$.push($dist, node.feature$)
        \If{$|H| > k$}
            \State $H$.pop()
        \EndIf
        \State $\tau \leftarrow \text{MaxDistance}(H)$
    \EndIf

    \If{$dist < node.median$}
        \State \Call{Search}{$node.left$}
        \If{$|H| < k$ \textbf{or} $|dist - node.median| < \tau$} \Call{Search}{$node.right$} \EndIf
    \Else
        \State \Call{Search}{$node.right$}
        \If{$|H| < k$ \textbf{or} $|dist - node.median| < \tau$} \Call{Search}{$node.left$} \EndIf
    \EndIf
\EndProcedure
\end{algorithmic}
\end{algorithm}

\paragraph{Searching the tree} Once built, the tree groups strongly related attributes near each other, so a $ k$-nearest-neighbor search only needs to visit a small part of the tree to find the attributes most strongly related to a query attribute $f_q$. Algorithm~\ref{alg:knn_search} performs this search with a branch-and-bound strategy: it keeps a running radius $\tau$, the distance to the current $k$-th best match, and skips any subtree whose attributes cannot possibly be closer than $\tau$.

Running this search for every attribute produces a sparse, asymmetric adjacency matrix $A$, where $A_{ij}$ is the dependency strength of attribute $f_j$ on $f_i$. Only the $k$ strongest dependencies per attribute are kept, giving an approximation of the full pairwise $Q$-matrix without ever computing all $n^2$ pairs directly.

\subsection{Experimental Setup}

\textbf{Datasets and preprocessing: }We evaluate the proposed framework on $14$ tabular datasets, comprising three simulated and eleven real-world datasets, with $8$ to $642$ attributes and approximately $100$ to $45000$ instances. The datasets were selected to cover diverse domains, sizes, and feature dimensionalities, enabling evaluation across both low- and high-dimensional settings. Table \ref{table:dataset_table} summarizes the characteristics of all datasets together with the number of extracted LDs and FDs. Before dependency extraction, missing values are removed, and categorical attributes are numerically encoded. For the \texttt{METABRIC} dataset, gene expression values are discretized into three quantile bins, mutation features are binarized, and only first-layer dependencies are extracted because of its high feature dimensionality. For validation, the extracted LDs are examined over the complete dataset to verify that they satisfy the adopted LD criterion, while the extracted FDs are compared with those returned by \textit{FDTool} across the evaluated dependency layers. For the simulated datasets, the explicitly designed LDs and FDs also serve as partial ground truth for validating dependency extraction, since the generated data may contain additional valid dependencies arising from transitive relationships or combinations of attributes.

\textbf{Parameter settings: }Unless stated otherwise, \textit{LDTool} uses a minimum dependency strength threshold of $\tau_s=0.8$ ($q\leq0.2$) to extract all FDs and strong LDs and an LD improvement threshold of $\tau_{LD}=0.2$. We set $\tau_{LD}=0.2$ because smaller values retain many additional LDs with only marginal improvements in dependency strength, whereas larger values do not further reduce the retained LDs. For \textit{HLDTool}, hypergraphs are constructed using $\tau_H=0.2$ with a maximum hyperedge size of $20$. We use $\tau_H=0.2$ as a default to enable consistent comparisons across datasets of different dimensions. The influence of $\tau_H$ is evaluated in Section \ref{results-hldtool}. 

\textbf{Implementation details: }The proposed framework was implemented in Python $3.11$ and executed on a server equipped with two NVIDIA RTX A100 GPUs ($80$ GB memory each), an AMD EPYC $7763$ CPU ($256$ cores), and $4$ TB RAM. The source code and preprocessed datasets are publicly available in \href{https://github.com/Chaithra-U/LDTool-and-HDTool}{GitHub} repository.

\section{Results}\label{sec: results} 

The experimental evaluation demonstrates the effectiveness, scalability, and computational efficiency of the proposed framework. We show that \textit{LDTool} extracts meaningful multi-attribute LDs, \textit{HLDTool} enables scalable discovery in high-dimensional feature spaces, and the proposed VP-tree approximation accelerates first-layer dependency extraction. Finally, we justify the selected parameter settings through sensitivity analysis.

\begin{figure*}[ht]
    \centering    \includegraphics[width=\textwidth]{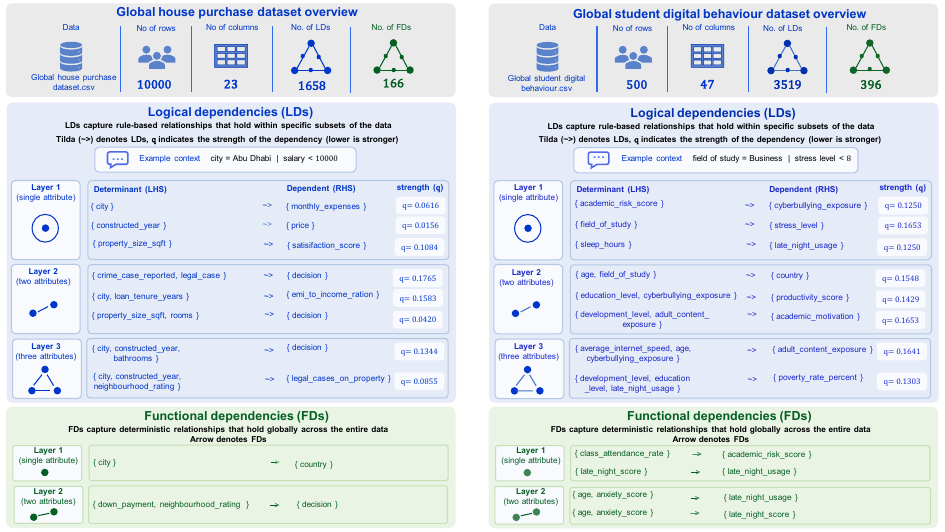}
    \caption{Extracted logical and functional dependencies from the \texttt{global house purchase} and \texttt{global student digital behavior} datasets using \textit{LDTool}. The extracted LDs capture meaningful relationships, such as those between city and salary and between field of study and stress level. Since FDs are a special case of LDs, \textit{LDTool} identifies both local logical patterns and globally valid dependencies. Lower $q$-values indicate stronger logical dependencies.}
    \label{figure:dependencies}
\end{figure*}

\textbf{\textit{LDTool} extracts meaningful multi-attribute logical dependencies from tabular datasets:}
We evaluated \textit{LDTool} on three simulated datasets and eleven publicly available real-world datasets (See Table \ref{table:dataset_table}) to analyze its ability to extract multi-attribute LDs. Dependencies were extracted up to the third layer for all datasets except the high-dimensional \texttt{METABRIC} and \texttt{Tic Insurance} datasets, where only the first- and second-layer dependencies were considered. The total number of extracted LDs for each dataset is summarized in Table \ref{table:dataset_table}, and the run time is recorded in Table \ref{table:time}, while representative outputs from two datasets are illustrated in Figure \ref{figure:dependencies}. 

The extracted dependencies reveal semantically interpretable relationships from the data. For example, in the \texttt{global student digital behavior} dataset (see Figure \ref{figure:dependencies}, to the right), the \textit{field of study} attribute is logically dependent on \textit{stress level}. Investigating the dataset, we find that when the field of study falls within the Business or STEM categories, stress levels are consistently lower (not above $8$ on a scale of $10$) than in the Medicine category. Similarly, in the same dataset, \textit{late-night usage} shows an LD with \textit{sleep hours}, where higher late-night usage corresponds to shorter sleep duration. 

In the \texttt{global house purchase} dataset (see Figure \ref{figure:dependencies}, to the left), \textit{city} and \textit{customer salary} are logically dependent, as customers located in Abu Dhabi consistently belong to higher salary ranges. Similarly, the combination of \textit{crime cases reported} and \textit{legal cases on property} is logically dependent on the \textit{purchase decision}. Properties associated with a legal case are consistently rejected, regardless of the number of reported crime cases. Furthermore, even in the absence of a legal case, properties with three or more reported crime cases are always rejected. These examples represent only a subset of the dependencies identified by \textit{LDTool}. Complete dependency outputs for all datasets are provided as supplementary \texttt{.txt} files in the \href{https://github.com/Chaithra-U/LDTool-and-HDTool/tree/main/dependencies}{GitHub} repository.

\begin{table}[ht]
\centering
\footnotesize
\setlength{\tabcolsep}{2.5pt}

\begin{tabular}{lrr}
\toprule
Dataset & Designed dependencies & Extracted by \textit{LDTool} \\
\midrule
Simulated\_data\_3 & 4  & 4  \\
Simulated\_data\_4 & 11 & 10 \\
Simulated\_data\_7 & 6  & 6  \\
\bottomrule
\end{tabular}

\caption{Validation of \textit{LDTool} using explicitly designed dependencies in the simulated datasets.}
\label{tab:simulated_validation}
\end{table}

In addition to the dependencies identified in the real-world datasets, the simulated datasets provide partial ground truth to validate the dependency-extraction capability of \textit{LDTool}. The reference set is considered partial because the generated data may contain additional valid dependencies arising from transitive relationships or from combinations of attributes not explicitly specified during data generation. Table \ref{tab:simulated_validation} compares the number of explicitly designed dependencies with those extracted by \textit{LDTool}. All designed dependencies were extracted for \texttt{Simulated\_data\_3} and \texttt{Simulated\_data\_7}, while $10$ of the $11$ designed dependencies were extracted for \texttt{Simulated\_data\_4}. The remaining LD was excluded because its $q$-score did not satisfy the selected strong-LD criterion ($q\leq0.2$).

\begin{figure*}[ht]
\centering

\subfloat[
Scalability of \textit{HLDTool}. With $\tau_H=0.2$, increasing the maximum hyperedge size recovers more LDs and FDs while increasing runtime.
\vspace{0.1cm}
\label{fig:hyperedge_size}
]{
    \includegraphics[width=0.95\textwidth]{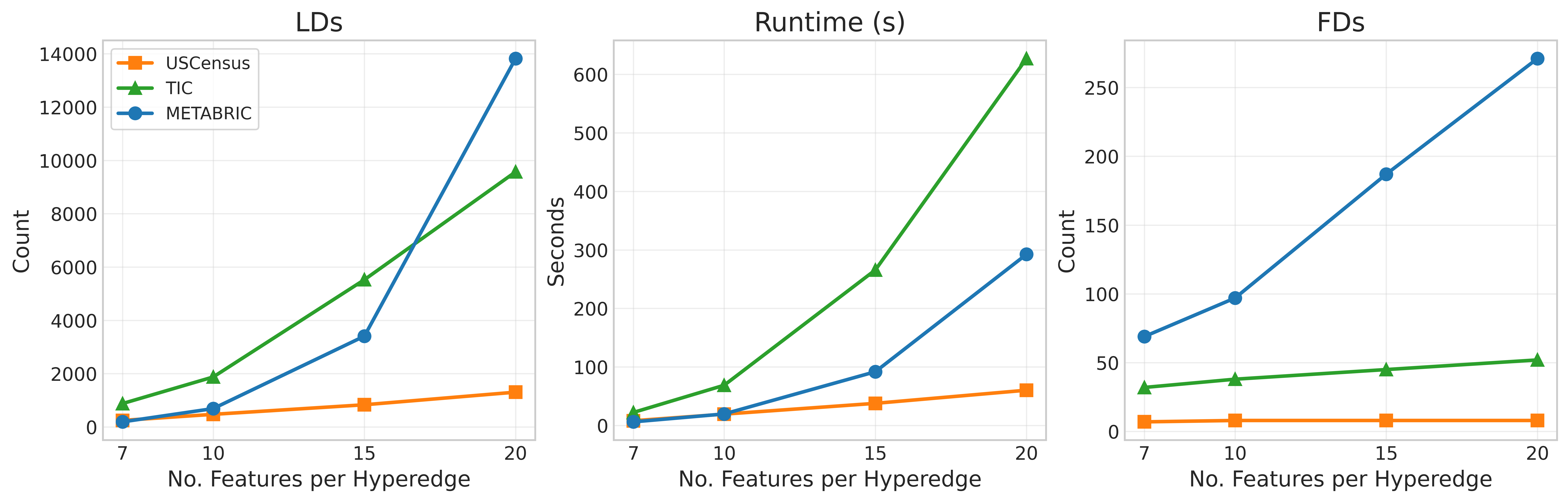}
}

\medskip

\subfloat[
Effect of the hypergraph construction threshold $\tau_H$. With the maximum hyperedge size fixed at $20$, increasing $\tau_H$ groups more attributes into hyperedges, increasing the number of recovered LDs and FDs while gradually increasing runtime.
\label{fig:hyperedge_threshold}
]{
    \includegraphics[width=0.95\textwidth]{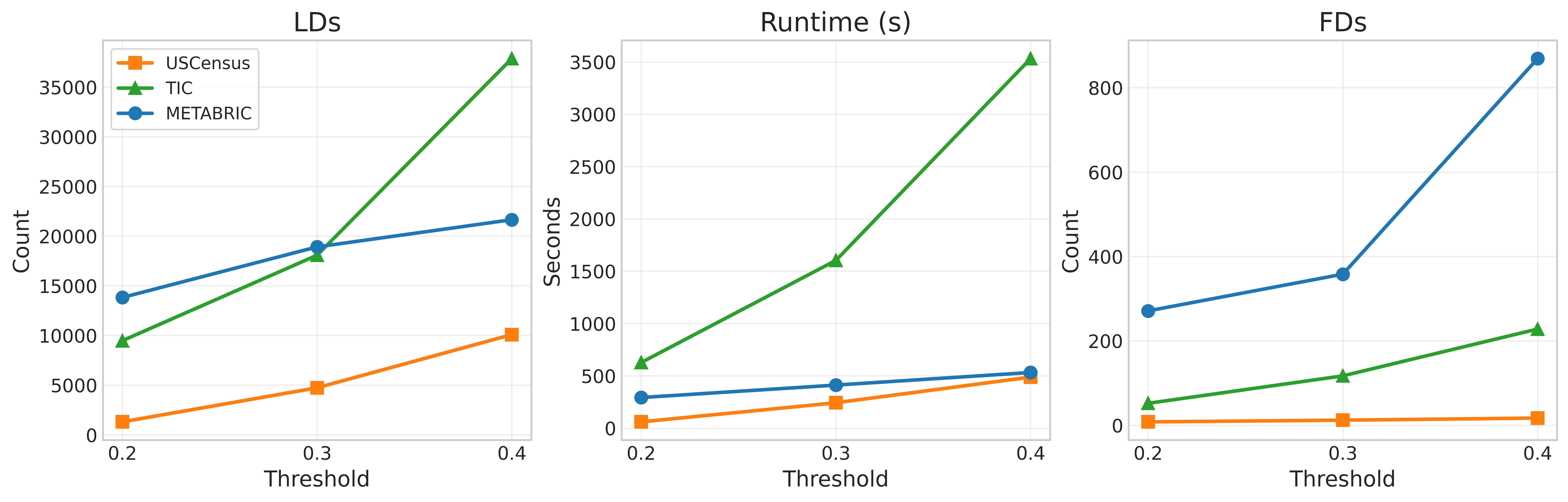}
}
\vspace{0.5cm}
\caption{Evaluation of \textit{HLDTool} on the \texttt{US Census}, \texttt{TIC Insurance}, and \texttt{METABRIC} datasets. (a) Scalability with respect to the maximum hyperedge size. (b) Effect of the hypergraph construction threshold $\tau_H$. Increasing either parameter recovers higher-order LDs and FDs, but at the cost of gradually increasing runtime. As exhaustive higher-order dependency extraction becomes computationally infeasible for high-dimensional datasets, \textit{HLDTool} provides a scalable and configurable alternative that enables practical dependency discovery.}
\label{fig:hldtool_scalability}
\end{figure*}

\textbf{\textit{HLDTool} enables scalable discovery of multi-attribute LDs and FDs in high-dimensional feature spaces:} \label{results-hldtool} Figure \ref{fig:hldtool_scalability}(a) demonstrates that \textit{HLDTool} provides a controllable trade-off between dependency coverage and computational cost across the \texttt{US Census} ($68$ features), \texttt{TIC Insurance} ($86$ features), and \texttt{METABRIC} ($642$ features) datasets. Increasing the maximum hyperedge size consistently recovers more LDs and FDs at the expense of increased runtime.

For the \texttt{US Census} dataset, exhaustive extraction remains feasible, enabling direct comparison with \textit{LDTool} and \textit{FDTool}. \textit{LDTool} extracts all third-layer dependencies ($224767$ LDs and $20$ FDs) in $306$ minutes. When restricted to FD extraction, \textit{LDTool} recovers the same $20$ FDs in $44$ minutes, compared with $60$ minutes for \textit{FDTool}. Using \textit{HLDTool} with a maximum hyperedge size of $20$ reduces the runtime to approximately one minute while recovering $1306$ LDs and $8$ FDs.

For the higher-dimensional \texttt{TIC Insurance} and \texttt{METABRIC} datasets, exhaustive third-layer extraction with \textit{LDTool} is computationally impractical, taking days or weeks. In contrast, \textit{HLDTool} discovers thousands of higher-order LDs and FDs within minutes by restricting the search to localized hyperedges. Since exhaustive higher-order extraction is infeasible for these datasets, the complete set of dependencies cannot be established. Nevertheless, the consistent trends across all three datasets demonstrate that \textit{HLDTool} enables practical discovery of higher-order dependencies in feature spaces where exhaustive search is computationally infeasible.

\begin{table}[ht]
\centering
\footnotesize
\setlength{\tabcolsep}{2.5pt}

\begin{tabular}{lrrrrrr}
\toprule
&
\multicolumn{3}{c}{\textbf{Brute-force}} &
\multicolumn{3}{c}{\textbf{VP-tree}} \\
\cmidrule(lr){2-4}
\cmidrule(lr){5-7}

\textbf{Dataset} &
Time &
LDs &
FDs &
Time &
LDs &
FDs \\
\midrule

Mushroom & 1.05s & 31 & 2 & 0.93s & 31 & 2\\
House purchase & 1.34s & 187 & 1 & 0.46s & 187 & 1\\
Consumer shopping & 1.43s &129&0&0.44s&129&0\\
Simulated\_data\_7 &1.69s&10&4&0.46s&10&4\\
Student behavior &3.00s&617&35&0.54s&436&34\\
US census &6.70s&110&4&6.30s&103&4\\
Tic insurance &22.28s&378&6&12.72s&334&6\\
METABRIC &452.6s&1512&7&40.8s&1450&7\\

\bottomrule
\end{tabular}

\caption{Comparison of brute-force and VP-tree computation of the $Q$-matrix for first-layer LD and FD extraction. The VP-tree substantially reduces runtime while recovering most dependencies.}
\label{tab:fast-tree-vs-brute-force}

\end{table}

Figure \ref{fig:hldtool_scalability}(b) further shows that increasing the hypergraph construction threshold $\tau_H$ progressively expands the search space by grouping more attributes into hyperedges, resulting in additional recovered LDs and FDs while gradually increasing runtime. Together, the maximum hyperedge size and $\tau_H$ provide complementary mechanisms for controlling the search space explored by \textit{HLDTool}. Larger hyperedges and higher values of $\tau_H$ prioritize dependency coverage, whereas smaller hyperedges and lower values of $ \tau_H$ prioritize computational efficiency. This flexibility allows \textit{HLDTool} to be adapted to datasets with different dimensionalities and computational budgets without modifying the underlying extraction algorithm.

\textbf{VP-tree-based neighborhood search accelerates first-layer dependency extraction:}
Computing first-layer LDs and FDs in \textit{LDTool} requires evaluating pairwise $q$-scores between all feature pairs, making the computation increasingly expensive as feature dimensionality grows. To accelerate this process, we employed the proposed VP-tree-based approximation to construct the dependency matrix and compared it with exhaustive $Q$-matrix computation (Table \ref{tab:fast-tree-vs-brute-force}).

Across all datasets, the VP-tree approach consistently reduced runtime while preserving most first-layer dependencies. For low- and medium-dimensional datasets, the recovered dependencies closely matched those obtained by exhaustive computation. For higher-dimensional datasets, including the \texttt{global student digital behavior} and \texttt{METABRIC} datasets, the VP-tree recovered over $70\%$ of the dependencies within a few seconds. In particular, for the \texttt{METABRIC} dataset, exhaustive computation required approximately seven minutes, whereas the VP-tree completed in seconds while recovering more than $95\%$ of the dependencies. These results demonstrate that the proposed VP-tree provides an efficient approximation for first-layer dependency extraction, making dependency matrix construction practical for medium- and high-dimensional datasets.

\textbf{Sensitivity analysis of LD extraction parameters:}
The number of extracted LDs in \textit{LDTool} and \textit{HLDTool} depends on the minimum dependency strength threshold $(\tau_s)$ and the LD improvement threshold $(\tau_{LD})$. Since dependency strength is defined as $1-q$, a threshold of $\tau_s=0.8$ (equivalent to $q \leq 0.2$) was selected to retain strong LDs while simultaneously capturing all FDs $(q=0)$. Lower values of $\tau_s$ resulted in a substantial increase in the number of extracted LDs, many of which corresponded to weaker dependency relationships.

The LD improvement threshold $(\tau_{LD})$ controls the retention of higher-order LDs during minimality filtering. Values of $\tau_{LD}<0.2$ produced a large increase in the number of extracted LDs, often yielding thousands of additional dependencies with only marginal improvements in dependency strength. Conversely, increasing $\tau_{LD}$ beyond $0.2$ did not further reduce the number of retained LDs across the evaluated datasets. Therefore, $\tau_{LD}=0.2$ was adopted, as it yielded a set of LDs without discarding additional informative dependencies.

\section{Discussion}

\textbf{Equivalent FD discovery with improved scalability in high-dimensional feature spaces:} 
Across all datasets, \textit{LDTool} extracted the same number of FDs as \textit{FDTool}, while requiring substantially lower runtime in high-dimensional feature spaces. Runtime comparisons are summarized in Table \ref{table:time}.

It is noteworthy that the \textit{FDTool} failed to extract a single FD from the high-dimensional \texttt{METABRIC} dataset. For datasets containing fewer than $25$ features, both approaches exhibited comparable execution times. However, as dimensionality increased, the computational advantage of the proposed framework became more pronounced. In particular, \textit{FDTool} failed to complete execution on the $642$-feature \texttt{METABRIC} dataset, even when restricted to first-layer dependency extraction, whereas \textit{LDTool} successfully extracted first-layer FDs in less than two minutes. \textit{LDTool} extracts LDs and FDs simultaneously in the same time it takes to extract FDs in low to moderate dimensional datasets.

\begin{table*}[!ht]
\centering
\footnotesize
\setlength{\tabcolsep}{2.5pt}

\begin{tabular}{lrrrr}
\toprule
\textbf{Dataset} &
\textbf{Features} &
\textbf{FDTool} &
\textbf{LDTool (FD)} &
\textbf{LDTool (FD+LD)} \\
\midrule

\href{https://github.com/Chaithra-U/LDTool-and-HDTool/blob/main/datasets/simulated_data_3.csv}
{Simulated\_data\_3}
& 8 & 0:00.084 & 0:00.492 & 0:00.881 \\

\href{https://github.com/Chaithra-U/LDTool-and-HDTool/blob/main/datasets/simulated_data_4.csv}
{Simulated\_data\_4}
& 15 & 0:00.364 & 0:01.766 & 0:09.088 \\

\href{https://www.kaggle.com/datasets/sahilislam007/shopping-trends-and-customer-behaviour-dataset}
{Shopping behavior}
& 15 & 0:02.536 & 0:05.186 & 0:15.424 \\

\href{https://archive.ics.uci.edu/dataset/2/adult}
{Adult}
& 15 & 0:04.173 & 0:23.036 & 0:36.305 \\

\href{https://archive.ics.uci.edu/dataset/468/online+shoppers+purchasing+intention+dataset}
{Online shopping}
& 18 & 0:05.919 & 0:26.273 & 0:45.920 \\

\href{https://codeocean.com/capsule/1269964/tree/v1/data/migraine.csv}
{Migraine}
& 20 & 0:02.029 & 0:10.035 & 0:37.046 \\

\href{https://archive.ics.uci.edu/dataset/73/mushroom}
{Mushroom}
& 22 & 0:00.974 & 0:21.824 & 1:20.733 \\

\href{https://www.kaggle.com/datasets/mohankrishnathalla/global-house-purchase-decision-dataset}
{Global house purchase}
& 24 & 3:00.159 & 1:07.611 & 2:41.143 \\

\href{https://www.kaggle.com/datasets/sohaibdevv/consumer-shopping-behavior-and-preference-study-2026}
{Consumer shopping trends 2026}
& 25 & 4:20.312 & 1:37.197 & 3:06.191 \\

\href{https://github.com/Chaithra-U/LDTool-and-HDTool/blob/main/datasets/simulated_data_7.csv}
{Simulated\_data\_7}
& 33 & 2:38.713 & 1:59.730 & 7:32.579 \\

\href{https://www.kaggle.com/datasets/nitikachandel95/student-social-media-impact-dataset?select=global_student_digital_behavior_dataset.csv}
{Student digital behavior}
& 47 & 0:22.114 & 1:02.043 & 79:35.056 \\

\href{https://archive.ics.uci.edu/dataset/116/us+census+data+1990}
{US Census}
& 68 & 60:01.108 & 44:44.357 & 306:30.402 \\

\href{https://archive.ics.uci.edu/dataset/125/insurance+company+benchmark+coil+2000}
{Tic Insurance}
& 86 & 110:21.278 & 92:14.199 & 3123:25.472 \\

\href{https://www.kaggle.com/datasets/raghadalharbi/breast-cancer-gene-expression-profiles-metabric}
{METABRIC}
& 642 & -- & 1:38.735 & 7:23.401 \\

\bottomrule
\end{tabular}

\caption{Runtime comparison between \textit{FDTool} and the proposed \textit{LDTool} for dependency extraction up to the first three layers across the same datasets, except for \texttt{METABRIC}, which is restricted to the first layer due to its high dimensionality. Runtime is reported as minutes: seconds (mm:ss.sss). While \textit{FDTool} extracts only FDs, \textit{LDTool} extracts both LDs and FDs. For smaller datasets, \textit{LDTool} extracts both dependency types in runtime comparable to \textit{FDTool}. When restricted to FD extraction, \textit{LDTool} requires less time, particularly for datasets with a larger number of features,
while identifying the same number of FDs.}
\label{table:time}

\end{table*}

\textbf{Importance of multi-attribute dependency discovery:} Many relationships in real-world tabular data arise from the combined influence of multiple attributes rather than individual feature pairs. The proposed framework extends dependency discovery beyond pairwise analysis by identifying multi-attribute LDs and FDs that capture such interactions. For example, in the \texttt{global house purchase} dataset, the purchase \textit{decision} depends jointly on the presence of \textit{legal cases} and the number of reported \textit{crime cases}, illustrating a relationship that cannot be fully characterized by either attribute alone. This ability to identify higher-order dependencies provides a richer characterization of the underlying structure of tabular data, supporting more informative exploratory analysis and interpretation.

\begin{figure}[ht!]
\centering

\begin{subfigure}[t]{0.48\linewidth}
    \centering
    \includegraphics[width=\linewidth,height=5cm,keepaspectratio]
    {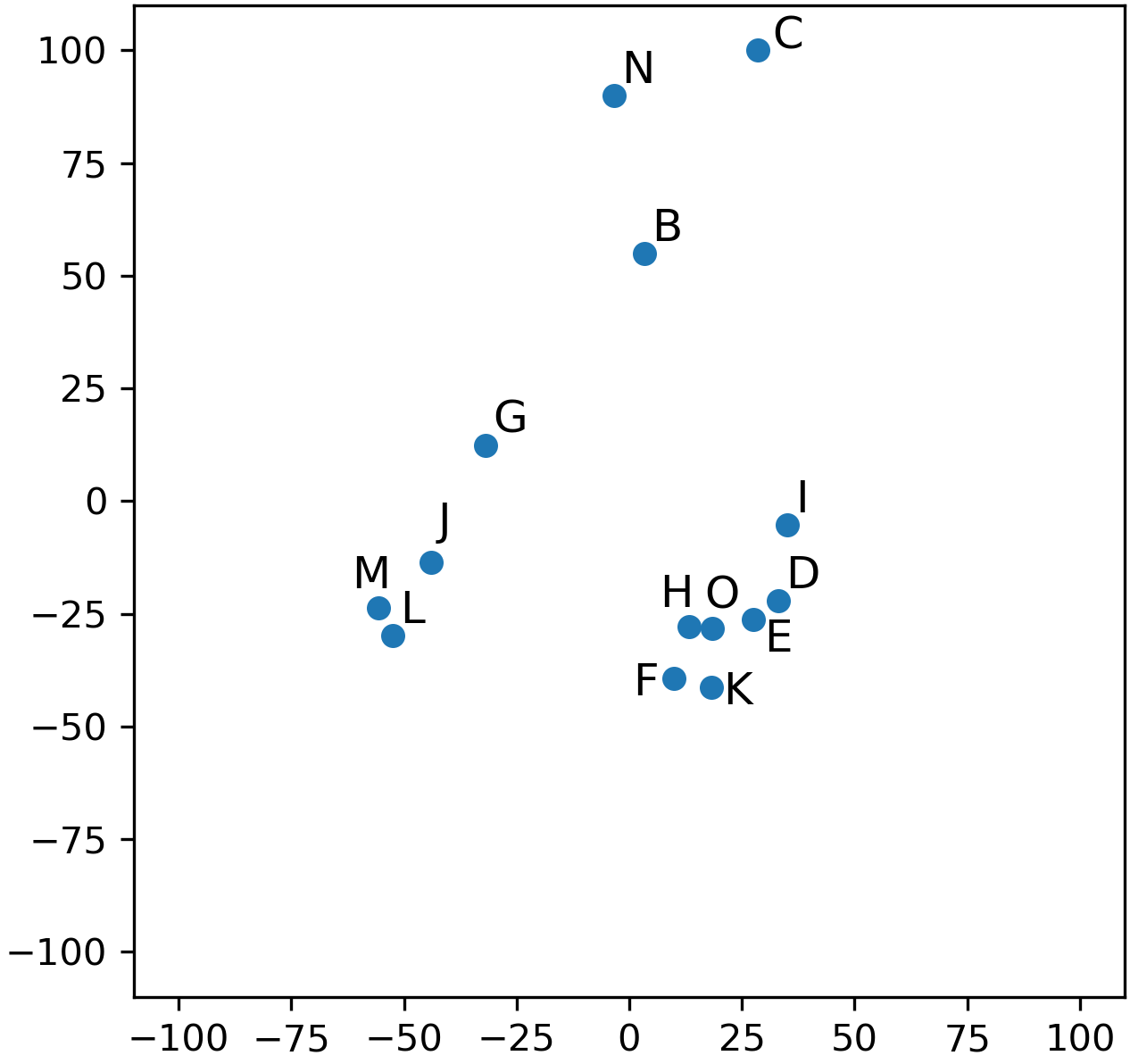}
    \caption{Force-directed visualization}
\end{subfigure}
\hfill
\begin{subfigure}[t]{0.48\linewidth}
    \centering
    \includegraphics[width=\linewidth,height=5cm,keepaspectratio]
    {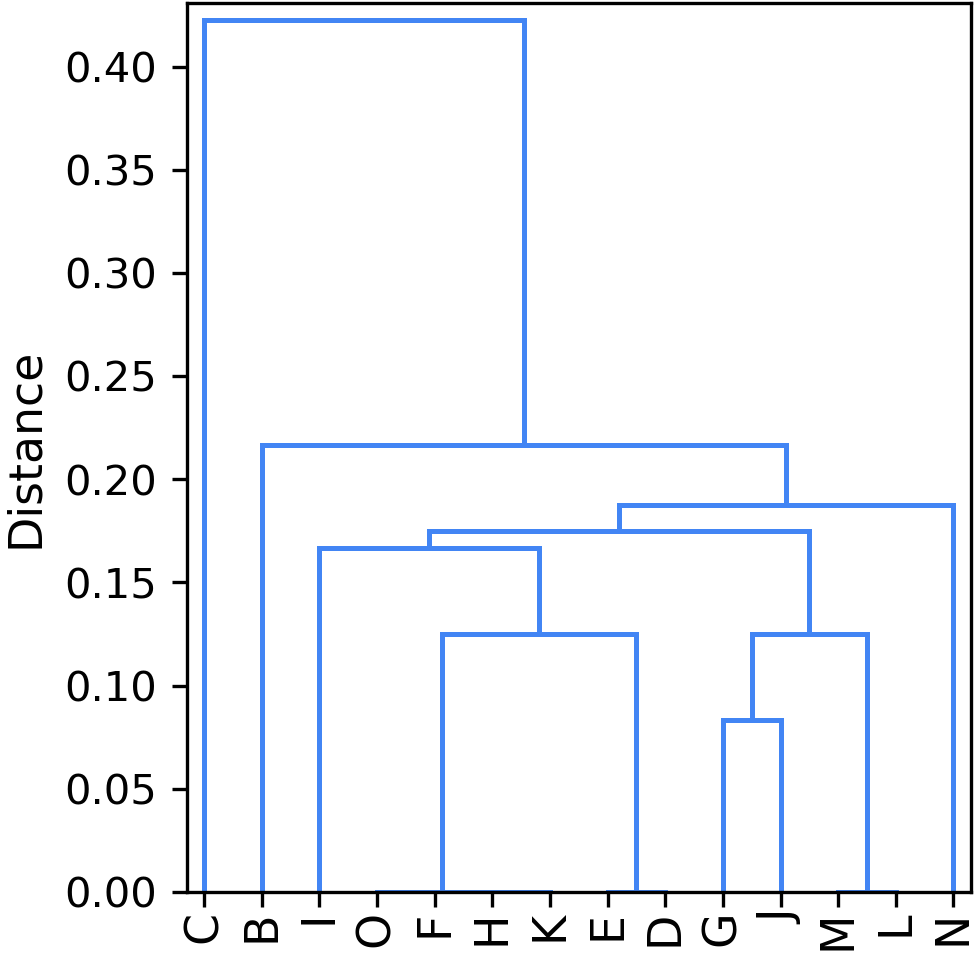}
    \caption{Dendrogram visualization}
\end{subfigure}

\vspace{1.8em}

\begin{subfigure}[t]{0.48\linewidth}
    \centering
    \includegraphics[width=\linewidth,height=5cm,keepaspectratio]
    {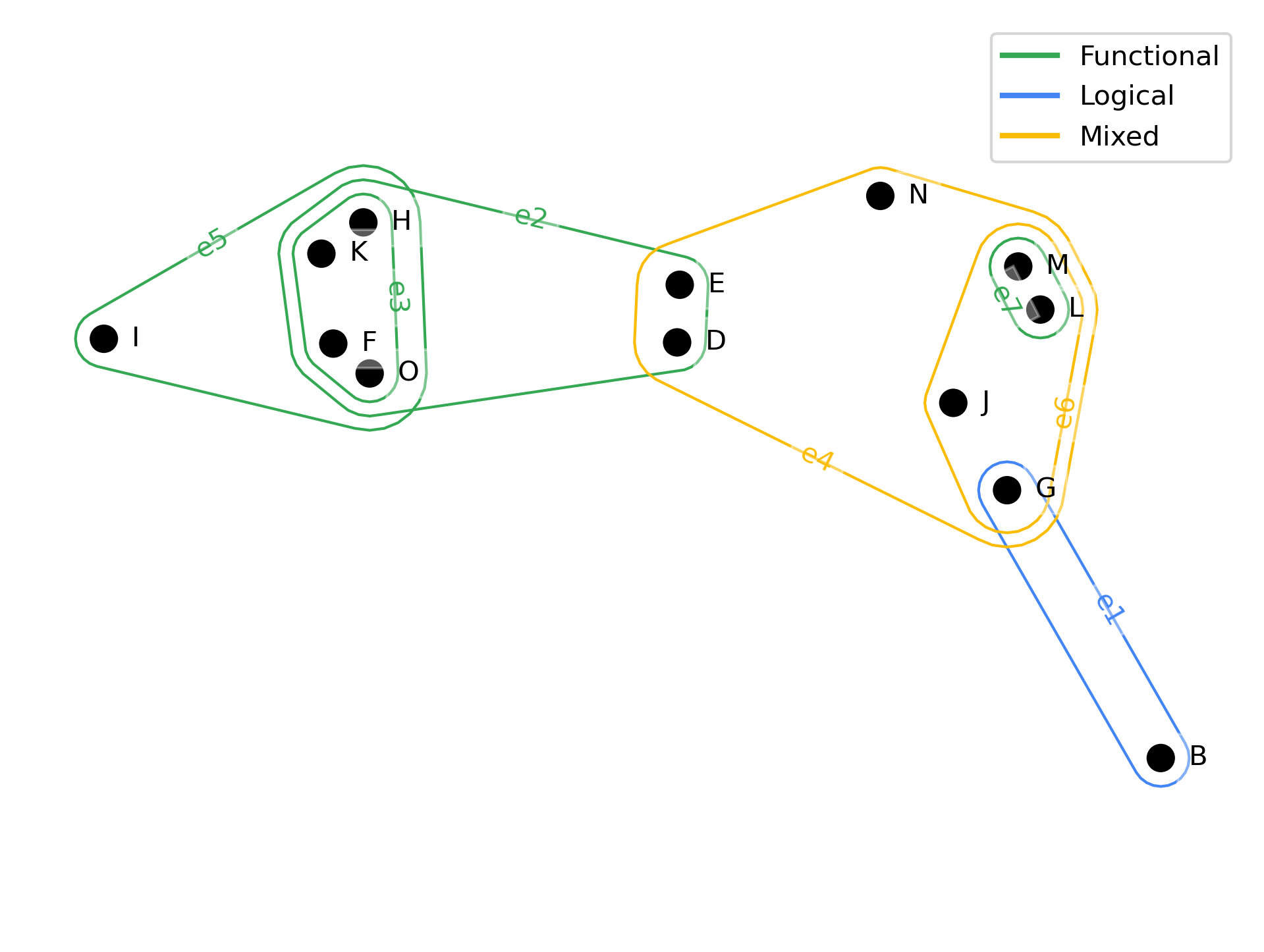}
    \caption{Hypergraph visualization}
\end{subfigure}
\hfill
\begin{subfigure}[t]{0.48\linewidth}
    \centering
    \includegraphics[
        width=0.90\linewidth,
        height=5cm,
        keepaspectratio
        ]
    {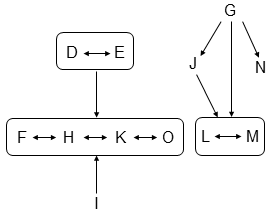}
    \caption{\textit{FDTool} extracted dependencies}
\end{subfigure}
\vspace{1.8em}
\caption{Comparison of force-directed, dendrogram, hypergraph, and manually constructed \textit{FDTool} visualizations for \texttt{simulated\_data\_4}. All visualizations consistently recover the same dependency groups. In the \textit{FDTool} graph, arrows denote FDs and double-headed arrows denote bidirectional (bijective) dependencies.}
\label{fig:visualization}
\end{figure}

\textbf{Multi-view visualizations reveal interpretable dependency structures:}
Figure \ref{fig:visualization} compares force-directed, dendrogram, and hypergraph visualizations generated from the pair-wise dependency scores with a manually constructed dependency diagram derived from \textit{FDTool}. 
Across all three visualization methods, structurally related attributes consistently form coherent groups that agree with the manually constructed dependency structure. The force-directed layout embeds attributes in two dimensions using the $q$-score as an interaction force, highlighting global connectivity among related attributes. The dendrogram applies hierarchical clustering to the $Q$-matrix, revealing nested groups of dependent attributes. The hypergraph represents multi-attribute dependencies via threshold-derived hyperedges, thereby making these relationships explicit. Together, these complementary visualizations provide an interpretable representation of multi-attribute dependency structures beyond textual rule lists. 

\textbf{Limitations and open research questions:}
While \textit{HLDTool} enables scalable discovery of multi-attribute LDs and FDs in high-dimensional feature spaces, assessing the completeness of the extracted dependency set remains challenging. For low- and moderate-dimensional datasets, the results can be compared with those obtained by \textit{LDTool} and \textit{FDTool} over the full feature space. In high-dimensional settings, however, exhaustive extraction of higher-order dependencies is computationally infeasible, making it difficult to establish a complete reference set. Furthermore, because \textit{HLDTool} restricts the search to feature groups defined by pairwise dependency-based hyperedges, multi-attribute dependencies involving features with weak pairwise relationships may be missed. This raises an important open question of how completeness can be assessed while retaining scalability in high-dimensional dependency discovery.

Beyond these methodological limitations, broader questions remain regarding the applicability of higher-order LDs and FDs. Their ability to reveal interpretable relationships in large-scale biological and biomedical datasets requires further investigation beyond the \texttt{METABRIC} dataset considered here. Another promising research question is whether LDs and FDs can complement existing measures for evaluating structural fidelity in synthetic tabular data and, more broadly, whether such dependencies can be incorporated into generative models to better preserve multi-attribute relationships.

\section{Conclusion}

This paper presents \textit{LDTool} and \textit{HLDTool}, a unified framework for extracting and visualizing multi-attribute logical and functional dependencies in tabular data. While \textit{LDTool} enables exhaustive LD and FD discovery up to a user-specified layer, \textit{HLDTool} improves scalability in high-dimensional feature spaces through hypergraph-guided search, with VP-tree-based approximation further accelerating pairwise dependency discovery. Experiments on simulated and real-world datasets demonstrate the extraction of multi-attribute LDs alongside FDs. When restricted to FD extraction, \textit{LDTool} recovers the same FDs as \textit{FDTool} for the evaluated dependency layers while requiring less runtime for datasets with larger feature spaces. The proposed graph, dendrogram, and hypergraph representations further support the interpretation of complex dependency structures. However, because \textit{HLDTool} restricts multi-attribute search using pairwise dependency information, dependencies among features with weak pairwise relationships may be missed, and assessing completeness remains challenging in high-dimensional settings.

Overall, the framework provides a practical and scalable approach for studying multi-attribute dependencies in tabular data, with potential applications to high-dimensional biological and biomedical datasets. Further research should investigate strategies for assessing dependency coverage when exhaustive extraction is computationally infeasible. Another promising direction is to examine whether LDs and FDs can complement existing measures of structural fidelity in synthetic tabular data and whether incorporating such dependencies into generative models can improve the preservation of multi-attribute relationships.

\section*{CRediT authorship contribution statement}
\textbf{Chaithra Umesh:} Conducted the experiments, analyzed the results, and wrote the first draft of the manuscript. \textbf{Arvind Lomrore:} Implemented the approximation algorithm experiment and contributed to the related work section. \textbf{Neethu D:} Conducted the literature survey and contributed to the related work section. \textbf{Kristian Schultz:} Discussed and reviewed the mathematical part of the manuscript. \textbf{Saptarshi Bej:} Conceptualized, reviewed, and supervised the manuscript writing and experiments. \textbf{Olaf Wolkenhauer:} Reviewed and supervised the manuscript writing and experiments.

\section*{Conflict of Interest}
The authors have no conflict of interest.

\section*{Availability of code and results}
We provided detailed Jupyter notebooks of our experiments on \href{https://github.com/Chaithra-U/LDTool-and-HDTool}{GitHub} to support transparency, reusability, and reproducibility.

\section*{Acknowledgment}
This work has been supported by the German Research Foundation (DFG) under Project No. 576429337, obtained for `Preserving logical and functional dependencies in synthetic clinical tabular data'.

\section*{Declaration of generative AI and AI-assisted technologies in the manuscript preparation process}
During the preparation of this work, the authors used ChatGPT (OpenAI) to generate a reference image to visualize the extracted dependencies. The generated image was used solely as a design reference, and the final figure (Figure \ref{figure:dependencies}) was manually created by the authors. No AI-assisted tools were used to generate the manuscript text, analyses, or scientific conclusions.

\bibliographystyle{elsarticle-num}
\bibliography{main}

\end{document}